\documentclass[conference]{IEEEtran}

\makeatletter

\def\ps@IEEEtitlepagestyle{%
  \def\@oddfoot{\mycopyrightnotice}%
  \def\@evenfoot{}%
}
\def\mycopyrightnotice{%
  {\footnotesize 979-8-3503-9591-4/24/\$31.00~\copyright~2024 IEEE\hfill}
  \gdef\mycopyrightnotice{}
}

\usepackage{blindtext}
\usepackage{eso-pic}
\IEEEoverridecommandlockouts
\usepackage{cite}
\usepackage{amsmath,amssymb,amsfonts}
\usepackage{algorithmic}
\usepackage{graphicx}
\usepackage{textcomp}
\usepackage{xcolor}
\usepackage{enumitem}
\usepackage{amsmath,epsfig}
\usepackage{amsmath}
\usepackage{amsfonts}
\usepackage{multirow}
\usepackage{multicol}
\usepackage{adjustbox}
\usepackage{mathtools}
\usepackage{amssymb}
\usepackage{caption}
\usepackage{subcaption}
\def\BibTeX{{\rm B\kern-.05em{\sc i\kern-.025em b}\kern-.08em
    T\kern-.1667em\lower.7ex\hbox{E}\kern-.125emX}}
    
\usepackage{eso-pic}
\newcommand\AtPageUpperMyright[1]{\AtPageUpperLeft{%
 \put(\LenToUnit{0.17\paperwidth},\LenToUnit{-2cm}){%
     \parbox{0.9\textwidth}{\raggedleft\fontsize{8}{11}\selectfont #1}}%
 }}%
\newcommand{\conf}[1]{%
\AddToShipoutPictureBG*{%
\AtPageUpperMyright{#1}
}
}    

\begin{document}

\def\x{{\mathbf x}}
\def\L{{\cal L}}
\def\OurMethod{Aligned Divergent Pathways}
\def\OurMethodAbrv{ADP}
\def\OurNorm{Dynamic Max-Deviance Adaptive Instance Normalization}
\def\OurNormAbrv{DyMAIN}
\def\OurMetric{Dimensional Consistency Metric Loss}
\def\OurMetricAbrv{DCML}
\def\OurPhasedCos{Phased Mixture-of-Cosines}
\def\OurPhasedCosAbrv{PMoC}

\title{\vspace*{1cm} \OurMethod{} for Omni-Domain Generalized Person Re-identification\\
}

\author{
\IEEEauthorblockN{Eugene P.W. Ang}
\IEEEauthorblockA{\textit{Rapid-Rich Object Search (ROSE) Lab} \\
\textit{Nanyang Technological University}\\
Singapore, Singapore \\
phuaywee001@e.ntu.edu.sg}
\and
\IEEEauthorblockN{Shan Lin}
\IEEEauthorblockA{\textit{Rapid-Rich Object Search (ROSE) Lab} \\
\textit{Nanyang Technological University}\\
Singapore, Singapore \\
shan.lin@ntu.edu.sg}
\and
\IEEEauthorblockN{Alex C. Kot}
\IEEEauthorblockA{\textit{Rapid-Rich Object Search (ROSE) Lab} \\
\textit{Nanyang Technological University}\\
Singapore, Singapore \\
eackot@ntu.edu.sg}


}

\maketitle
\conf{\textit{  IV. International Conference on Electrical, Computer and Energy Technologies (ICECET 2024) \\ 
25-27 July 2024, Sydney-Australia}}
\begin{abstract}
Person Re-identification (Person ReID) has advanced significantly in fully supervised and domain generalized Person ReID. However, methods developed for one task domain transfer poorly to the other. An ideal Person ReID method should be effective regardless of the number of domains involved in training or testing. Furthermore, given training data from the target domain, it should perform at least as well as state-of-the-art (SOTA) fully supervised Person ReID methods. We call this paradigm \textit{Omni-Domain Generalization} Person ReID, referred to as ODG-ReID, and propose a way to achieve this by expanding compatible backbone architectures into multiple diverse pathways. Our method, \OurMethod{} (\OurMethodAbrv{}), first converts a base architecture into a multi-branch structure by copying the tail of the original backbone. We design our module \OurNorm{} (\OurNormAbrv{}) that encourages learning of generalized features that are robust to omni-domain directions and apply \OurNormAbrv{} to the branches of \OurMethodAbrv{}. Our proposed \OurPhasedCos{} (\OurPhasedCosAbrv{}) coordinates a mix of stable and turbulent learning rate schedules among branches for further diversified learning. Finally, we realign the feature space between branches with our proposed \OurMetric{} (\OurMetricAbrv{}). \OurMethodAbrv{} outperforms the state-of-the-art (SOTA) results for multi-source domain generalization and supervised ReID within the same domain. Furthermore, our method demonstrates improvement on a wide range of single-source domain generalization benchmarks, achieving Omni-Domain Generalization over Person ReID tasks.
\end{abstract}



\section{Introduction}
Person Re-identification (ReID) matches the image of a query person with other images captured in other nonoverlapping cameras. The simplest fully supervised single-domain ReID tasks train and test within a single domain of cameras, whereas in domain generalized ReID (DG-ReID), training and test sets come from mutually exclusive camera domains. The best single-domain ReID methods perform significantly worse when tested on other unseen domains, but the same problem exists in the other direction for methods specialized in domain generalization: overspecialization to the generalization task also affects performance in the simple single domain setting. Illustrated in Table~\ref{tab:test}, a baseline single domain ResNet-50 method~\cite{StrongBaseline} performs much worse than specialized DG-ReID methods when evaluated across domains. However, even when provided with samples from the test domain for training, these state-of-the-art (SOTA) methods specialized in DG-ReID \textit{severely underperform a simple baseline}, suggesting that these methods are overly specialized for the generalization task. An ideal Person ReID method that has truly learnt to generalize should do as well, if not better, when trained with data from the test domain. We call this paradigm \textit{Omni Domain Generalization Person Re-identification}, referred to as ODG-ReID. In this work, we introduce our solution, \OurMethod{} (\OurMethodAbrv{}), that is agnostic to the above domain configuration changes and can achieve ODG-ReID.

\begin{table}[ht]
\center
\captionsetup{justification=centering}
\caption{The performance of single-domain methods degrades when testing on other datasets, while cross-domain generalization methods cannot maintain the same performance on the original training dataset. Dataset abbreviations: M is Market-1501~\cite{Market-1501}, D is DukeMTMC-reID~\cite{DukeMTMC-reID}, C is CUHK03~\cite{CUHK03}, and MS is MSMT17$\_$V2~\cite{MSMT17}. Dataset(s) on the left side of the arrow are used in training and evaluation is performed on the dataset on the right. All results are in mAP and the best result for each benchmark is bold.}
\begin{adjustbox}{width=1\linewidth}
\begin{tabular}{c|c|cc|cc}
\hline
\multirow{2}{*}{Setting} & \multirow{2}{*}{Methods} & \multicolumn{2}{c|}{Single Domain Train} & \multicolumn{2}{c}{Mult-Domain Train} \\ \cline{3-6} 
& & \multicolumn{1}{c|}{\textbf{D} → \textbf{D}}  & M → \textbf{D} & \multicolumn{1}{c|}{C+\textbf{D}+M → \textbf{D}} & C+M+MS → \textbf{D} \\ \hline
Normal & Baseline (BoT)~\cite{StrongBaseline} & \multicolumn{1}{c|}{\textbf{76.3}} &  16.1   & \multicolumn{1}{c|}{\textbf{76.3}}         &     38.6      \\ \hline
\multirow{3}{*}{Domain Generalized} & DualNorm~\cite{DualNorm}       & \multicolumn{1}{c|}{67.7} &   23.9  & \multicolumn{1}{c|}{76.3}         &  40.5         \\
                         & META~\cite{META-DGReID}           & \multicolumn{1}{c|}{46.7} &  18.3   & \multicolumn{1}{c|}{60.6}       &  42.7         \\
                         & ACL~\cite{ACL-DGReID}            & \multicolumn{1}{c|}{49.9} &  22.2   & \multicolumn{1}{c|}{72.9}     & \textbf{53.4}          \\ 
                         & SIL~\cite{StyleInterleaved} & \multicolumn{1}{c|}{61.5} & \textbf{25.1} & \multicolumn{1}{c|}{65.6} &  47.0 \\ \hline
\end{tabular}%
\end{adjustbox}
\label{tab:test}
\end{table}

\noindent Given a standard backbone architecture such as a ViT~\cite{ViT} or ResNet~\cite{ResNet}, our framework expands the backbone with branches by making multiple tail copies. We add our proposed normalization method, \OurNorm{} (\OurNormAbrv{}) to each branch, which infuses \textit{maximal style variation} into each data batch, cross-pollinating styles between pairs that are most different in style from one another. This encourages the learning of generalized features that are robust to omni-domain generalization settings. 

Next, our \OurPhasedCos{} (\OurPhasedCosAbrv{}) learning rate schedules coordinate a diverse mix of schedules among the branches, from stable schedules with small learning rates and long cycles, to turbulent schedules with large learning rates and short cycles. 

Finally, we propose a novel metric learning loss, \OurMetric{} (\OurMetricAbrv{}), that enforces alignment between the inter-pathway feature dimensions. \OurMethodAbrv{} outperforms SOTA multi-source domain generalization and single-domain supervised ReID benchmarks, demonstrating omni-domain generalization. Furthermore, we show in our ablation studies that the components of \OurMethodAbrv{} improve performance on many single-source domain generalization benchmarks.

Our contributions are summarized as follows:
\setlist{nolistsep}
\begin{itemize}[noitemsep]
    \item We investigate Omni-Domain Person Re-identification, referred to as ODG-ReID, demonstrating that single domain methods cross domains ineffectively, while DG-ReID methods perform poorly in settings where test domain samples are provided for training. ODG-ReID is a crucial capability in real-world applications but, to the best of our knowledge, is rarely considered.
    \item We propose a novel and highly adaptable framework, \OurMethodAbrv{}, that can be applied on popular backbone architectures such as ViTs and ResNets.
    \item \OurMethodAbrv{} outperforms SOTA Person ReID performance in multi-source generalization and single-domain benchmarks. Furthermore, we demonstrate improvements on a wide range of single-source domain generalization tasks.
\end{itemize}

\section{Related Work}

\textbf{Single-Domain Person Re-identification}
Single-domain ReID SOTA methods are mostly based on deep convolutional neural networks. Early deep learning based work usually formulated Person ReID as a verification problem and proposed Siamese architectures \cite{CUHK03, Huang2019Multi-PseudoRe-Identification, Lin2017End-to-EndRe-Identification}. Then, verification-driven triplet architectures, such as \cite{TriNet, Quadruplet} overtook Siamese structures to add robustness. Zheng et al.~\cite{IDE} first proposed the widely adopted ID-Discriminative embedding (IDE) to train models pretrained on ImageNet \cite{ImageNet} to classify person IDs. More recent approaches \cite{MGN,StrongBaseline,AGW-pami21reidsurvey} combine classification and verification losses. 

\noindent\textbf{Domain-Generalized Person ReID}: DG-ReID aims to learn a model that can generalize well to unseen target domains without using any target domain data for training. Instance normalization (IN) was used to normalize style variation~\cite{IBN-Net} and is particularly successful in DG-ReID~\cite{DualNorm}. Adaptive IN (AdaIN)~\cite{AdaIN} extends this concept by mixing style features from an image with the content features of a target image. Our proposed normalization, \OurNormAbrv{}, extends AdaIN to select style sources based on maximal deviance. MMFA-AAE~\cite{MMFA-AAE} used a domain adversarial learning approach to remove domain-specific features. QAConv~\cite{QAConv} and M$^3$L~\cite{M3L} used meta-learning frameworks coupled with memory bank strategies. DEX~\cite{DEX} proposed a domain-based implicit semantic augmentation loss function to learn domain invariant features. ACL~\cite{ACL-DGReID} proposed a module to separately process domain invariant and domain specific features, plugging the module to replace selected convolutional blocks. RaMoE~\cite{RaMoE-Dai2021} and META~\cite{META-DGReID} deploy a mixture of experts to specialize to each domain. Style Interleaved Learning (SIL)~\cite{StyleInterleaved} is a framework with a regular forward/backward pass, with one additional forward pass that employs style interleaving on features to update class centroids. The recent development of Vision Transformers (ViT)~\cite{ViT} birthed a new generation of DG-ReID methods using ViT backbones. For instance, Part-Aware Transformer~\cite{ni2023part} learns locally similar features shared across different IDs, further using the part-guided information for self-distillation.

\begin{figure}[ht]
\centering
\includegraphics[width=0.9\columnwidth]{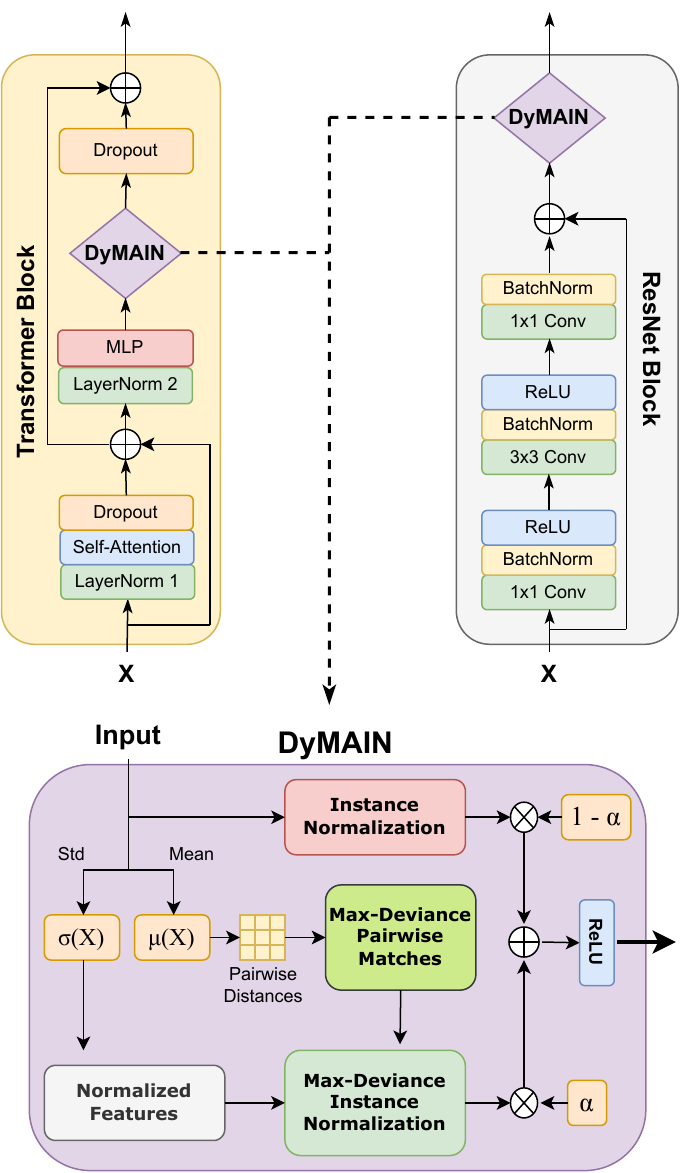} 
\caption{\OurNormAbrv{} can be applied on Transformer and ResNet blocks. The internal structure of our \OurNormAbrv{} module is illustrated in detail in the flow-diagram at the bottom of the figure.}
\label{fig-dynamic-max-deviance-adaptive-instance-norm}
\end{figure}
\section{Methodology}
\subsection{Preliminaries}
We are given a dataset of person images with identity labels $D=\{x_i,y_i\}_1^n$. For ViT backbones, we sample the image $x_i$ in a grid, linearize them into a sequence of image patches which we embed into a sequence of features. We then append a class token to the front of the sequence and add positional encodings to preserve the embedding order. We notate a batch of such sequences as $X \in \mathbb{R}^{N, S, d}$, where $N$ is batch size, $S$ is the sequence length (class token plus number of image patches) and $d$ is the embedding dimension. $X$ forms the input to a series of transformer encoder blocks that further specialize the sequence of encodings for relevant tasks. At the end of the encoder blocks, the class token is used for training or inference. For ResNets, we follow well established standards~\cite{StrongBaseline}.

\subsection{\OurNorm{}}
\OurNormAbrv{} is illustrated in Fig~\ref{fig-dynamic-max-deviance-adaptive-instance-norm}. Given a batch of inputs $X$ we compute the feature-wise means $\mu(X)$ and standard deviations $\sigma(X)$, which we use to normalize the inputs: $\Bar{X_i} = \frac{X_i - \mu(X_i)}{\sqrt{\sigma_i^2(X_i) + \epsilon}}$ where $\epsilon$ is a small value for numerical stability. Further, we compute the pairwise distances between the feature means $\mu(X)$ and use it to match batch samples: each sample $X_i$ is paired with its most distant counterpart $X_{M_i}$ to derive a set of matchings $\{i,M_i\}_{i=1}^N$. For our Max-Deviance Adaptive Instance Normalization (MAIN), we mix the style feature of sample $X_{M_i}$ into its paired content feature $X_i$: $\text{IN}_{MA}(X_i) = (\frac{\Bar{X_i}}{\sigma(X_{M_i})} + \mu(X_{M_i})) \gamma + \beta$ where $\gamma$ and $\beta$ are learnable affine parameters. \OurNormAbrv{} combines MAIN with regular Instance Normalization as $\text{IN}_{DyMA}(X) = \sum_{i=1}^{N} \mathbf{\alpha}  \text{IN}_{MA}(X_i) + (1-\mathbf{\alpha})\text{IN}(X_i)$, using a learnable parameter $\mathbf{\alpha} \in \mathbb{R}^d$ that combines the outputs of both normalizations along the embedding dimension. 


\begin{figure}[ht]
     \centering
     \begin{subfigure}[b]{0.5\textwidth}
         \centering
         \includegraphics[trim={0.1cm 0.2cm 1.5cm 0.9cm},clip, width=\textwidth]{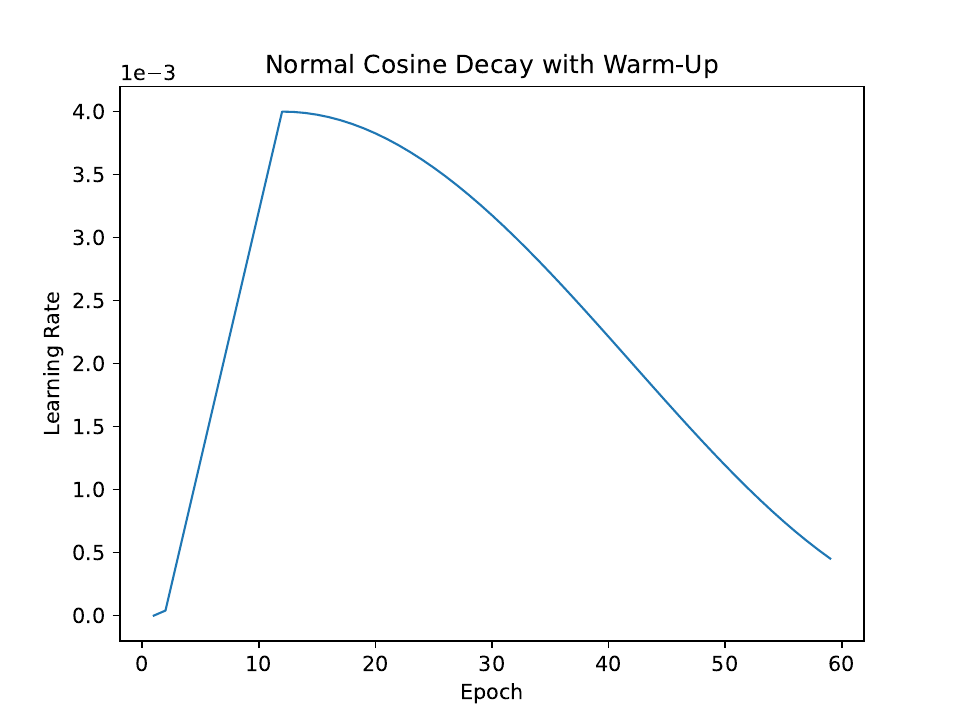}
         \caption{}
         \label{fig:cos-schedule-vanilla}
     \end{subfigure}
     \begin{subfigure}[b]{0.5\textwidth}
         \centering
         \includegraphics[trim={0.1cm 0.2cm 1.5cm 0.9cm},clip, width=\textwidth]{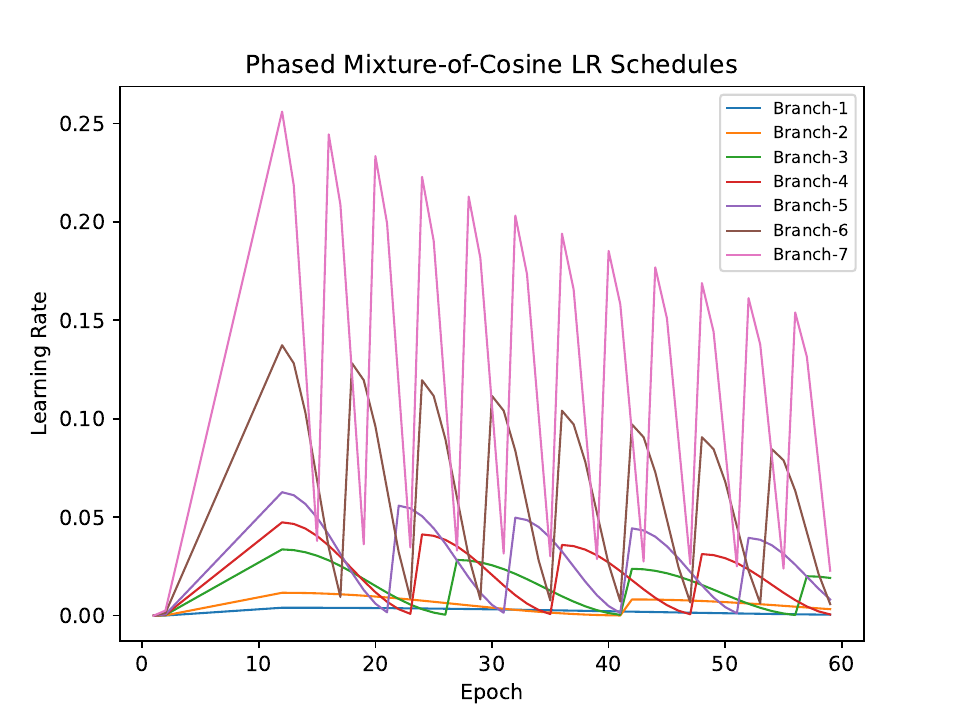}
         \caption{}
         \label{fig:prime-phased-cos-lr-schedule}
     \end{subfigure}
    \caption{Comparing a standard cosine learning rate schedule (a) with our ensemble of Phased Mixture-of-Cosine (PMoC) schedulers. The mix of turbulent and stable schedules in PMoC encourages learning diverse features among different branches.}
    \label{fig:compare-cos-schedules}
\end{figure}

\subsection{\OurPhasedCos{}}
While cosine learning rate schedulers are effective in the training of deep neural networks, using the same scheduler in all branches overly correlates their performance as all go through the same stages of learning at the same time. We apply different schedules to each branch so that they explore different feature spaces at each point in time, collectively providing the model with divergent opinions. Fig~\ref{fig:cos-schedule-vanilla} illustrates a normal learning rate schedule. We deploy a mix of turbulent schedules with larger learning rates but shorter periods and stable schedules with smaller learning rates and longer periods. Formally, given total epochs $T$, minimum learning rate $\eta_{min}$, learning rate power factor $\gamma$, target decay factor $\lambda$, each branch $k$ with period $p_k$ epochs trains for $c_k = \lfloor \frac{T}{p_k} \rfloor$ cycles with a base learning rate of $\eta_k = (c_k)^{\gamma}\eta_{min}$. After each cycle, we decay the learning rate by a factor of $\lambda^{\frac{1}{c_k}}$. Fig~\ref{fig:prime-phased-cos-lr-schedule} illustrates example schedules that run multiple cycles with learning rate decay between cycles.

\begin{figure}[ht]
\centering
\includegraphics[width=0.7\columnwidth]{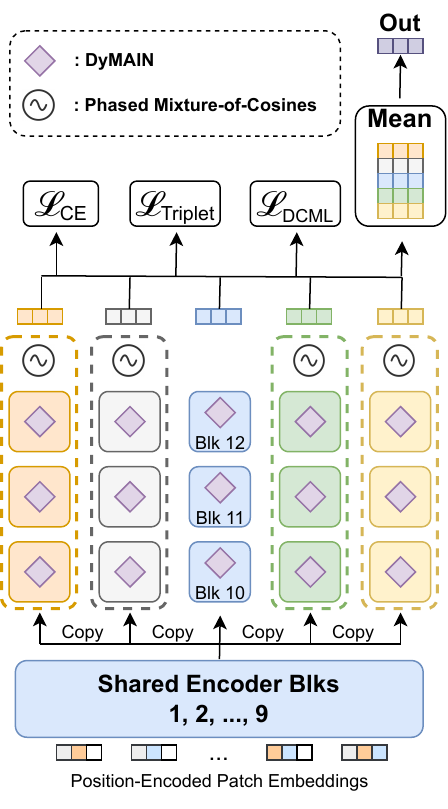} 
\caption{A summary of our method, \OurMethodAbrv{}. Image inputs are processed into sequences of patches and passed through a backbone model, such as a Vision Transformer. The final blocks of the backbone are cloned to make a self-ensemble of several branches. Our own normalization module DyMAIN is added to all branch blocks, and each branch specialized with a different cosine learning rate schedule via Phased Mixture-of-Cosines. A dimensional consistency metric loss (DCML) is imposed to further align the features from different branches.}
\label{fig-onsen-big}
\end{figure}

\subsection{\OurMethod{}}
Figure~\ref{fig-onsen-big} illustrates our method, \OurMethodAbrv{}. The final blocks of the backbone spawn multiple branches that share the earlier layers for a more parameter-efficient architecture. Each branch is specialized with \OurNormAbrv{}, allowing them to learn features with maximum style variation. Each branch explores different embedding space-time combinations due to the coordination of unique learning rate schedules from our \OurPhasedCosAbrv{} scheduler. We apply \OurMetricAbrv{} loss to encourage the features among sibling branches to be consistent and aligned. At the end of the inference pass, we combine all branch features together by taking the mean along the feature dimension.

\subsection{\OurMetric{}}
The diverse learning rate schedules coordinated by \OurPhasedCosAbrv{} results in a larger exploration space between branches, but a balance needs to be achieved between exploration and consistency. We propose our \OurMetricAbrv{} loss to keep inter-branch distances small and also encourage features to be more aligned in feature dimension. Given $\{X_i := f_i(X)\}_{i=1}^k$, a set of sibling feature outputs from $k$ branches (each denoted by a function $f_i$), we compute the Chebyshev distance between any pair of features $(x,y)$: $D_{\infty}(x, y) = \max_{i}(| x_i - y_i |)$. Dimensional consistency is encouraged by reducing the \textit{maximum component difference} between each pair of inter-branch features, resulting in our \OurMetricAbrv{} loss function:
\begin{equation}
    \L{}_{\OurMetricAbrv{}}(X) = \frac{1}{\binom{k}{2}} \sum_{(i,j) \in \binom{[1..k]}{2}} D_{\infty}(X_i,X_j)
\end{equation}
Other metrics such as Manhattan or Euclidean distances densely penalize deviations between all the dimensions of two input vectors. Such metrics degrade the final performance of the model. $\L_{\OurMetricAbrv{}}$, on the other hand, only penalizes component with the maximum absolute deviation and its sparsity helps with learning of generalized and aligned sibling features. 

\subsection{Loss Functions}
For a batch of input samples $X$ with labels $y$, \OurMethodAbrv{} generates output features from $k$ branches $\{f_i(X)\}_{i=1}^k$. We apply a per-branch cross-entropy loss over the person identity labels: 
\begin{equation}
\L_{CE}^{b}(X,y) = -\log \frac{\exp{\mathbf{w}_y^b \cdot f_b(X)}}{\sum_{j=1}^{C}\exp{\mathbf{w}_j^b \cdot f_b(X)}} \; ,  
\end{equation}
where $\mathbf{w}^b$ are branch classifier weights, and $C$ is the total number of classes. We also apply triplet losses per-branch: 
\begin{equation}
\L_{Tri}^b(X) = \sum_{(a,p,n) \in T(f_b(X))} [d_{a,p} - d_{a,n} + m]_+ \; ,
\end{equation}
where $d$ is a distance metric, $m \in \mathbb{R}$ is a margin parameter and $a,p,n$ are anchor, positive and negative triplets mined from $f_b(X)$ using a triplet collator $T$ such as hard-example mining~\cite{TriNet}. Adding our \OurMetricAbrv{}, the overall loss is a weighted sum of the three individual losses balanced by hyperparameters $0 \leq w_1, w_2, w_3$:
\footnotesize
\begin{equation}
    L_{\OurMethodAbrv{}}(X) = \sum_{b=1}^{k} \left[ w_1 \L_{CE}^{b}(X) + w_2 \L_{Tri}^{b}(X) \right] + w_3\L_{\OurMetricAbrv{}}(X)
\end{equation}
\normalsize

\section{Experiments}
\label{sec:experiment-settings}
\noindent\textbf{Person ReID Datasets and Multi-Source Domain Generalization Benchmarks}
\noindent For multi-source DG-ReID benchmarks we adopt a leave-one-out evaluation where out of four datasets, Market-1501 (M), DukeMTMC-reID (D), CUHK03 (C) and MSMT17$\_$V2 (MS), three are used for training and one for testing, e.g. D+M+MS means that we train on D, M and MS and test on C. For CUHK03 we use the new partition, following~\cite{DEX}. Only the training sets of the source domains are used in training. 

\noindent\textbf{Single Domain ReID Benchmarks} The single domain benchmarks use the same ReID datasets (M, D, C and MS). In supervised ReID, we use the same dataset for training and testing. For single-source DG-ReID, we use the training split of a dataset for training and evaluate on the test split of another dataset. For CUHK03, to align our results those reported by SOTA single-domain methods, we use the classic split with detected bounding boxes.

\begin{table*}[ht]
\center
\captionsetup{justification=centering}
\caption{Results on the modern DG-ReID benchmarks. For the methods with the dagger symbol $\dag$, we evaluated the official open source implementation on this benchmark. \textbf{Bold} numbers are the best, while \underline{underlined} numbers are second. Ours (R50) is our method with the ResNet-50 backbone, while Ours (ViT) is with the Vision Transformer backbone.}
\resizebox{1.7\columnwidth}{!}{
\begin{tabular}{l|cc|cc|cc|cc}
\hline
\multirow{2}{*}{Method} &
  \multicolumn{2}{c|}{C+D+MS→M} &
  \multicolumn{2}{c|}{C+M+MS→D} &
  \multicolumn{2}{c|}{C+D+M→MS} &
  \multicolumn{2}{c}{D+M+MS→C} \\ \cline{2-9} 
                      & Rank-1 & mAP  & Rank-1 & mAP  & Rank-1 & mAP  & Rank-1 & mAP  \\ \hline
DualNorm~\cite{DualNorm} $\dag$   & 78.9   & 52.3 & 68.5   & 51.7 & 37.9   & 15.4 & 28.0   & 27.6 \\
QAConv~\cite{QAConv}     & 67.7   & 35.6 & 66.1   & 47.1 & 24.3   & 7.5  & 23.5   & 21.0 \\
OSNet-IBN~\cite{OSNet} $\dag$  & 73.4   & 45.1 & 61.5   & 42.3 & 35.7   & 13.7 & 20.9   & 20.9 \\
OSNet-AIN~\cite{OSNet} $\dag$  & 74.2   & 47.4 & 62.7   & 44.5 & 37.9   & 14.8 & 22.4   & 22.4 \\
M$^3$L~\cite{M3L}     & 75.9   & 50.2 & 69.2   & 51.1 & 36.9   & 14.7 & 33.1   & 32.1 \\
DEX~\cite{DEX}        & 81.5   & 55.2 & 73.7   & 55.0 & 43.5   & 18.7 & 36.7   & 33.8 \\
SIL~\cite{StyleInterleaved} $\dag$     & 79.2   & 52.8 & 68.3   & 47.0 &   36.9   & 13.8 & 29.5   & 28.9 \\
META~\cite{META-DGReID} $\dag$  & 66.7   & 44.6 & 61.8   & 42.7 & 32.4   & 13.1 & 21.3   & 21.6 \\
ACL~\cite{ACL-DGReID} $\dag$     & 86.1   & \underline{63.1} & 71.7   & 53.4 & 47.2   & 19.4 & 35.5   & 34.6 \\
Ours (R50)  & \underline{86.3}   & 62.0 & \underline{76.5}   & \underline{58.9} & \underline{48.3}   & 21.2 & \underline{38.4}   & \underline{37.6} \\ \hline
PAT~\cite{ACL-DGReID}    & 75.2   & 51.7 & 71.8   & 56.5 & 45.6   & \underline{21.6} & 31.1   & 31.5 \\
Ours (ViT) & \textbf{86.6}   & \textbf{64.5} & \textbf{77.4}   & \textbf{62.4} & \textbf{53.4}   & \textbf{25.7} & \textbf{41.6}   & \textbf{40.7} \\ \hline
\end{tabular}%
}
\label{tab:dg-reid-modern}
\end{table*}

\begin{table*}[h]
\vspace{1em}
\captionsetup{justification=centering}
\caption{Results on the Single Domain ReID Benchmarks. For the methods with the dagger symbol $\dag$, we evaluated the official open source implementation on this benchmark. \textbf{Bold} numbers are the best, while \underline{underlined} numbers are second. Ours (R50) is our method with the ResNet-50 backbone, while Ours (ViT) is with the Vision Transformer backbone.}
\center
\resizebox{2.1\columnwidth}{!}{
\begin{tabular}{cc|ccc|cccc|ccc}
\hline
\multicolumn{2}{c|}{\multirow{2}{*}{\textbf{Method}}}                                       & 
\multicolumn{3}{c|}{\textbf{Single-Domain}} & \multicolumn{4}{c|}{\textbf{Domain Generalization}} & \multicolumn{3}{c}{\textbf{Omni-Domain Generalization}}    \\
\multicolumn{2}{c|}{}                                                                       & 
BoT $\dag$ & MGN $\dag$ & AGW & DualNorm $\dag$ & SIL $\dag$ & META $\dag$ & ACL $\dag$ & OSNet-AIN $\dag$ & Ours (R50) & Ours (ViT) \\ \hline
\multirow{2}{*}{\textbf{Market-1501}} 
& mAP & 80.6 & 85.3 & \underline{88.2} & 76.4 & 71.9 & 47.2 & 73.1 & 78.5 & 87.6 & \textbf{88.5} \\
& R1  & 92.3 & 94.5 & \textbf{95.3} & 91.9 & 88.9 & 75.1 & 88.0 & 92.5 & 95.0 & \underline{95.1} \\ \hline
\multirow{2}{*}{\textbf{DukeMTMC-reID}}                                               
& mAP & 63.0 & 76.3 & \textbf{79.6} & 67.7 & 61.5 & 46.7 & 49.9 & 69.9 & 78.6 & \underline{79.5} \\
& R1  & 78.9 & 86.9 & 89.0 & 83.8 & 79.9 & 73.1 & 71.5 & 84.7 & \underline{89.5} & \underline{89.1} \\ \hline
\multirow{2}{*}{\textbf{CUHK03}} 
& mAP & 88.5 & 87.6 & 62.0 & 67.2 & 78.1 & 56.0 & 79.8 & 70.0 & \textbf{89.9} & \underline{89.7} \\
& R1  & 90.5 & 91.1 & 63.6 & 68.9 & 82.5 & 60.5 & 83.2 & 74.9 & \textbf{93.0} & \underline{92.8} \\ \hline
\multirow{2}{*}{\textbf{MSMT17 (V2)}} 
& mAP & 49.2 & 50.5 & 49.3 & 43.9 & 37.2 & 37.1 & 27.7 & 42.7 & \underline{61.0} & \textbf{62.0} \\
& R1  & 74.7 & 74.8 & 68.3 & 74.4 & 66.2 & 69.1 & 57.0 & 71.5 & \underline{83.8} & \textbf{85.0}\\ \hline
\end{tabular}
}
\label{tab:reid-single-domain}
\end{table*}

\noindent We evaluate all methods with mean average precision (mAP) and Rank-1.
\subsection{Implementation Details}
\label{sec:implementation-details}
We apply \OurMethodAbrv{} on two backbones, ViT-b-16~\cite{ViT} and ResNet-50~\cite{ResNet}, both with weights pretrained on ImageNet~\cite{ImageNet}. Input images are resized into 384x128 pixels. For ViT, patches of 16x16 pixels are sampled from the image in strides of 12, then linearized and position-encoded into a sequence of $\mathbb{R}^{768}$ embeddings. For both backbones we clone the last four blocks/bottlenecks to make each branch, creating a total of k=7 branches. \OurNormAbrv{} is applied to all blocks of each branch. Our cross-entropy, triplet and \OurMetricAbrv{} losses have weights of $w_1=1, w_2=1, w_3=0.01$. We use a SGD optimizer with a base learning rate of $\eta=0.004$ and momentum $0.9$. The main branch uses a cosine learning rate schedule, warming up linearly over ten epochs starting from $0.01\eta$ and attenuating over the course of training in one cycle, down to a minimum learning rate of $0.002\eta$. For \OurPhasedCosAbrv{}, we set $\eta_{min}=0.004, \gamma=1.806, \lambda=0.5$ and the branches cycle in periods of $p_{1...k}=[120,60,30,24,20,15,12]$ epochs respectively. We use a batch size of 64 and apply auto-augmentation~\cite{AutoAugment} with $p=0.1$. For each encoder block, we apply dropout with $p=0.1$. We train for a total of 120 epochs ($T=120$).

\subsection{Results on Cross Domain Generalization ReID}
Table~\ref{tab:dg-reid-modern} presents a comparison between our \OurMethodAbrv{} and other DG-ReID methods on the modern leave-one-out benchmarks. \OurMethodAbrv{} improves the domain generalization performance of the ViT backbone, outperforming recent SOTA methods by a considerable margin - our method surpasses the previous SOTA by 5.9\% in mAP for C+M+MS $\rightarrow$ D, by 4.1\% in mAP and 6.2\% in Rank-1 for C+D+M $\rightarrow$ MS and by 6.1\% in mAP and 4.9\% in Rank-1 for D+M+MS $\rightarrow$ C. \OurMethodAbrv{} also improves the capability of the ResNet-50, surpassing nearly all other similar methods based on the same backbone.

\subsection{Results on Single Domain ReID benchmarks}
Table~\ref{tab:reid-single-domain} compares \OurMethodAbrv{} against SOTA DG-ReID and single domain ReID methods. SOTA DG-ReID methods perform worse on single domain tasks because of over-specialization to domain generalization tasks. In contrast, \OurMethodAbrv{} reaches SOTA-level performance in all four single domain ReID benchmarks, even outperforming previous SOTA methods by 11.5\% in mAP and by 10.2\% in Rank-1 for the MSMT17 benchmark.

\subsection{Ablation Studies}
We conduct ablation experiments to find the best hyperparameter configurations of each of its components. For brevity, we only report results of \OurMethodAbrv{} on the ViT backbone. Through these experiments, we show that the components improve a wide range of single-source DG-ReID benchmarks.

\subsubsection{Ablation over application of \OurNormAbrv{}} Table~\ref{tab:ablation-dymain-blocks} shows how \OurNormAbrv{} improves domain generalization for selected benchmarks MS $\rightarrow$ M and M $\rightarrow$ C. Starting from a vanilla ViT control, we progressively add \OurNormAbrv{} to an increasing number of ending transformer encoder blocks. Performance improves steadily by nearly 2\% Rank-1 and mAP when we apply \OurNormAbrv{} on up to four of the ending blocks. We observe that adding \OurNormAbrv{} to more than the last four blocks begins to degrade performance.

\begin{table}[h]
\captionsetup{justification=centering}
\caption{Ablation over application of \OurNormAbrv{}. We experiment on a single-branch base model and gradually increase the number of ending blocks that we apply \OurNormAbrv{} to.}
\center
\resizebox{0.9\columnwidth}{!}{
\begin{tabular}{c|cc|cc}
\hline
\multicolumn{1}{c|}{\multirow{2}{*}{\textbf{Blocks Applied}}} & \multicolumn{2}{c|}{\textbf{MS $\rightarrow$ M}} & \multicolumn{2}{c}{\textbf{M $\rightarrow$ C}} \\
\multicolumn{1}{c|}{} & \textbf{mAP} & \textbf{R1} & \textbf{mAP} & \textbf{R1}     \\ \hline
\textbf{None} & 46.6 & 72.5 & 26.6 & 25.4  \\
\textbf{1} & 47.5 & 73.2 & 26.9 & 25.9            \\
\textbf{2} & 47.7 & 73.6 & 26.8 & 25.7            \\
\textbf{3} & 47.6 & 72.7 & 26.9 & 26.2            \\
\textbf{4} & \textbf{48.4} & \textbf{73.9} & \textbf{27.6} & \textbf{27.1}            \\
\textbf{5} & 47.9 & 73.2 & 26.5 & 25.4        \\  \hline   
\end{tabular}
}
\label{tab:ablation-dymain-blocks}
\end{table}
\vspace{1em}

\subsubsection{Comparing learning rate schedules} We compare our \OurPhasedCosAbrv{} (with implementation details as described in Subsection~\ref{sec:implementation-details}) against a stable learning rate schedule (Stable) where all branches apply the same cosine schedule with $\eta = 0.004$ over a single period. To ensure that a higher base learning rate is not the cause behind the improvement, we compare these two experiments against a learning rate schedule where all branches apply a high base learning rate of $\eta = 0.256$ (Turbulent). Table~\ref{tab:ablation-compare-lr-schedules} shows that diversity of learning rate schedules result in significant performance gains of around 2\% on the single-source domain generalization benchmarks MS $\rightarrow$ C and M $\rightarrow$ D.

\begin{table}[h]
\captionsetup{justification=centering}
\caption{Comparing \OurPhasedCosAbrv{} with Stable and Turbulent LR-Schedules. Mixing stable and turbulent schedules reaps the most benefit through diversity of learning among different branches.}
\center
\resizebox{1\columnwidth}{!}{
\begin{tabular}{l|cc|cc}
\hline
\multicolumn{1}{c|}{\multirow{2}{*}{\textbf{LR Regime}}} & \multicolumn{2}{c|}{\textbf{MS $\rightarrow$ C}} & \multicolumn{2}{c}{\textbf{M $\rightarrow$ D}} \\
\multicolumn{1}{c|}{} & \textbf{mAP} & \textbf{R1} & \textbf{mAP} & \textbf{R1}     \\ \hline
Stable ($\eta=0.004$) & 29.5 & 29.2 & 50.3 & 67.8            \\
Turbulent ($\eta=0.256$) & 27.6 & 26.9 & 50.2 & 68.1            \\
\textbf{\OurPhasedCosAbrv{} ($0.004 \leq \eta \leq 0.256$)} & \textbf{30.7} & \textbf{31.0} & \textbf{51.9} & \textbf{69.7}     \\ \hline      
\end{tabular}
}
\label{tab:ablation-compare-lr-schedules}
\end{table}
\vspace{1em}

\subsubsection{Ablation over weight of \OurMetricAbrv{}} Table~\ref{tab:ablation-metric-weights} compares the effects of different weights of \OurMetricAbrv{} over the benchmarks C $\rightarrow$ MS and C $\rightarrow$ M. Setting a loss weight of $w_3 = 0.01$ yields the most benefit over the control ($w_3 = 0$). With \OurMetricAbrv{}, we observe a performance improvement of around 1-2\% in both mAP and Rank-1.

\begin{table}[h]
\center
\captionsetup{justification=centering}
\caption{Ablation over weight of \OurMetricAbrv{}.}
\resizebox{0.9\columnwidth}{!}{
\begin{tabular}{l|cc|cc}
\hline
\multicolumn{1}{c|}{\multirow{2}{*}{\textbf{\OurMetricAbrv{} weight ($\lambda_3$)}}} & \multicolumn{2}{c|}{\textbf{C $\rightarrow$ MS}} & \multicolumn{2}{c}{\textbf{C $\rightarrow$ M}} \\
\multicolumn{1}{c|}{} & \textbf{mAP} & \textbf{R1} & \textbf{mAP} & \textbf{R1}     \\ \hline
\textbf{$w_3=0$} & 16.5 & 39.5 & 52.4 & 74.7  \\
\textbf{$w_3=0.1$} & 16.8 & 41.0 & 53.4 & 76.0            \\
\textbf{$w_3=0.01$} & \textbf{17.1} & \textbf{41.5} & \textbf{54.1} & \textbf{76.4}           \\ 
\textbf{$w_3=0.001$} & 16.4 & 39.9 & 51.8 & 74.4            \\
\hline
\end{tabular}
}
\label{tab:ablation-metric-weights}
\end{table}

\subsubsection{Components of \OurMethodAbrv{}} Putting the components together, Table~\ref{tab:ablation-components} presents the effects of incrementally adding these components of \OurMethodAbrv{}, demonstrating the gradual improvements contributed by each component.

\begin{table}[h]
\captionsetup{justification=centering}
\caption{Ablation over components of \OurMethodAbrv{}.}
\center
\resizebox{1\columnwidth}{!}{
\begin{tabular}{ccc|cc|cc}
\hline
\multicolumn{3}{c|}{\textbf{\OurMethodAbrv{} Components}}   & \multicolumn{2}{c|}{\textbf{C+D+M → MS}} & \multicolumn{2}{c}{\textbf{MSMT17}} \\
\textbf{\OurNormAbrv{}} & \textbf{\OurMetricAbrv{}} & \textbf{\OurPhasedCosAbrv{}} & \textbf{mAP}        & \textbf{R1}        & \textbf{mAP}      & \textbf{R1}     \\ \hline
 & & & 20.8 & 44.7 & 56.8 & 79.8  \\
\checkmark & & & 21.7 & 45.8 & 59.8 & 81.7 \\
 & \checkmark & & 23.0 & 50.5 & 60.1 & 81.5 \\
 & & \checkmark & 24.2 & 52.1 & 59.5 & 82.2 \\
\checkmark & \checkmark & & 24.8 & 50.6 & 60.2 & 83.6 \\
\checkmark & \checkmark & \checkmark & \textbf{25.7} & \textbf{53.4} & \textbf{62.0} & \textbf{85.0} \\ \hline
\end{tabular}
}
\label{tab:ablation-components}
\end{table}

\section*{Conclusion}
In this paper we investigated the omni-domain generalization capabilities of Person ReID methods and found that methods specialized in one task domain transfer poorly to other task domains. Although this was broadly within expectations, it was concerning to note that this performance degradation is observed \textit{even if the transfer task is of a simpler setting than the original task}. For example, SOTA multi-source domain generalization methods trained with the full data from the target domain still underperform simple fully supervised baselines. This suggests that such methods are too specialized for their task and do not succeed in varied domain settings. We proposed a novel framework, \OurMethodAbrv{}, to alleviate this overfit. Our method expands a standard backbone architecture into multiple branches, maximizes style differences in the data by applying our proposed normalization method \OurNorm{}, diversifies per-branch training regimes using our \OurPhasedCos{} learning rate schedules, and aligns interbranch features with our \OurMetric{}. \OurMethodAbrv{} outperforms recent SOTA methods in DG-ReID as well as in single domain ReID, demonstrating omni-domain generalization and taking a step towards real-world application.

\section*{Acknowledgements}
This work was supported by the Defence Science and Technology Agency (DSTA) Postgraduate Scholarship, of which Eugene P.W. Ang is a recipient. It was carried out at the Rapid-Rich Object Search (ROSE) Lab at the Nanyang Technological University, Singapore.

\bibliographystyle{ieeetr}

\end{document}